\documentclass[runningheads]{llncs}

\usepackage{eccv}

\usepackage{eccvabbrv}

\usepackage{graphicx}
\usepackage{booktabs}
\usepackage{todonotes}
\usepackage{multirow}
\usepackage{amssymb}%
\usepackage{pifont}%
\newcommand{\cmark}{\ding{51}}%
\newcommand{\xmark}{\ding{55}}%

\usepackage[accsupp]{axessibility}  %

\usepackage{hyperref}

\usepackage{orcidlink}

\begin{document}

\title{Level Up Your Tutorials: \\ VLMs for Game Tutorials Quality Assessment} 
\titlerunning{Level Up Your Tutorials}

\author{
Daniele Rege Cambrin\inst{1}\orcidlink{0000-0002-5067-2118} 
\and
Gabriele Scaffidi Militone \inst{1}\orcidlink{0009-0002-2462-8778} 
\and
Luca Colomba\inst{1}\orcidlink{0000-0003-2911-4522}
\and
Giovanni Malnati\inst{1}\orcidlink{0000-0002-6798-2761}
\and
Daniele Apiletti\inst{1}\orcidlink{0000-0003-0538-9775}
\and
Paolo Garza\inst{1}\orcidlink{0000-0002-1263-7522}
}

\authorrunning{D. Rege Cambrin et al.}
\institute{Politecnico di Torino, Turin, Italy\\
\email{\{daniele.regecambrin,gabriele.scaffidi,luca.colomba \\
giovanni.malnati,daniele.apiletti,paolo.garza\}@polito.it}
}

\maketitle
\begin{abstract}

Designing effective game tutorials is crucial for a smooth learning curve for new players, especially in games with many rules and complex core mechanics. Evaluating the effectiveness of these tutorials usually requires multiple iterations with testers who have no prior knowledge of the game.
Recent Vision-Language Models (VLMs) have demonstrated significant capabilities in understanding and interpreting visual content. VLMs can analyze images, provide detailed insights, and answer questions about their content. They can recognize objects, actions, and contexts in visual data, making them valuable tools for various applications, including automated game testing.
In this work, we propose an automated game-testing solution to evaluate the quality of game tutorials. Our approach leverages VLMs to analyze frames from video game tutorials, answer relevant questions to simulate human perception, and provide feedback. This feedback is compared with expected results to identify confusing or problematic scenes and highlight potential errors for developers. In addition, we publish complete tutorial videos and annotated frames from different game versions used in our tests.
This solution reduces the need for extensive manual testing, especially by speeding up and simplifying the initial development stages of the tutorial to improve the final game experience.
\end{abstract}

\begin{figure}
    \centering
    \includegraphics[width=\linewidth]{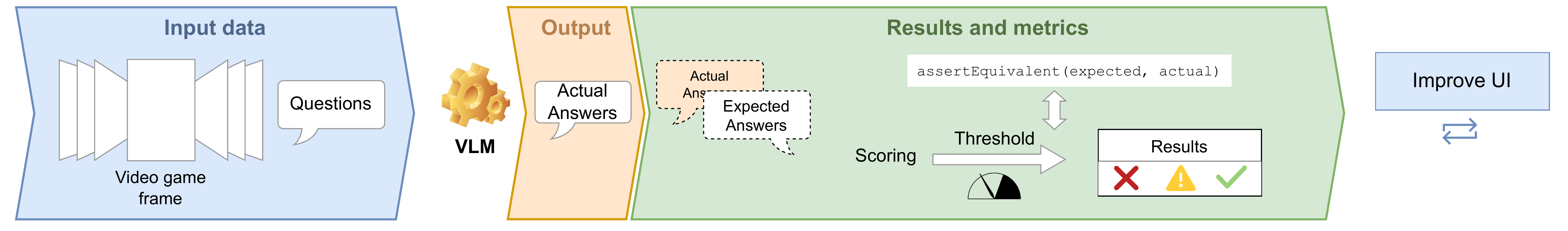}
    \caption{Proposed framework. VLM is asked to answer questions about tutorial frames. Actual answers by the VLM are then compared with the expected answers provided by the developers to provide a quality score. The score informs about possible areas of improvement that can be used to improve the final User Interaction (UI).}
    \label{fig:framework}
\end{figure}

\section{Introduction}

A crucial component of this initial experience is the game tutorial. Effective tutorials are essential for ensuring that players can quickly learn the fundamental mechanics and objectives of the game, thereby enhancing their overall enjoyment and engagement. This is particularly important for games with unique or complex core mechanics~\cite{tutorial_relevance}. However, designing and evaluating effective game tutorials traditionally involves multiple iterations with human testers who have no prior knowledge of the game. This process is time-consuming and resource-intensive and prone to inconsistencies and delays in identifying critical issues~\cite{videogameTester}.

Recent Vision-Language Models (VLMs) demonstrated remarkable capabilities in understanding and interpreting visual content~\cite{internvl,gpt4}. They can analyze images, provide detailed insights, and answer questions about their content, recognizing objects, actions, and contexts within visual data. These capabilities make VLMs valuable tools for various applications, including automated game testing.

In this work, we propose an innovative automated game-testing solution that leverages VLMs to evaluate the quality of game tutorials. Our approach involves analyzing frames from video game tutorials and using VLMs to answer relevant questions, simulating human perception. This method provides immediate feedback, which is then compared to expected outcomes defined by the developers. By identifying confusing or problematic scenes and highlighting potential bugs, our solution helps developers refine and improve the tutorials, ensuring a smoother learning curve for players.

Our solution involves extracting significant frames from game tutorials and labeling them with questions that a developer would ask a human tester. These questions, along with their expected answers, serve as a ground truth to evaluate the VLMs' answers. By comparing the VLMs' responses to the expected answers, we can determine the clarity and comprehensibility of the tutorials as shown in~\Cref{fig:framework}. This process is analogous to writing automated software tests, aiming to identify and correct issues early in the development cycle.

Automating the testing process enables developers to focus on refining game mechanics and delivering a better player experience. Compared to traditional game testing, which relies on iterative human feedback, this solution speeds up the evaluation process, allowing for more frequent and thorough assessments without the need for extensive human resources. Moreover, unlike human testers, VLMs provide consistent feedback through each iteration, and they can be easily scaled to different games and tutorial formats.

Our contributions can be summarized as follows:
\begin{enumerate}
    \item We propose a new solution to automatically evaluate the clarity of tutorials in games;
    \item We release the frame tutorials annotated by developers in two different versions of the game and the associated videos and the code at \url{https://github.com/DarthReca/level-up-your-tutorials};
    \item We benchmarked various open-source and closed-source state-of-the-art models for the proposed task
\end{enumerate}

The rest of this article is structured as follows.
In~\Cref{sec:related} the related works are presented.~\Cref{sec:dataset} describes the game, the tutorials and the dataset collection process adopted in the experimental section. Instead,~\Cref{sec:methodology} introduces the proposed solution, whose experimental results are presented in~\Cref{sec:results}.~\Cref{sec:conclusion} draws conclusions and provides suggestions for future work.

\section{Related Works}
\label{sec:related}

This section presents contributions from two relevant areas: Visual Language Models and Machine Learning for game testing.

\subsection{Vision Language Models}
\label{subsec:vlm}

In recent years, Vision-Language Models (VLMs) have gained significant attention within the computer vision and natural language processing communities, particularly for their ability to integrate visual and textual information. Key advancements in this domain have been driven by models such as CLIP~\cite{clip} and ALIGN~\cite{align_paper}, which leverage large-scale datasets to learn joint representations of images and texts. Recent advancements have seen the creation of Flamingo~\cite{flamingo}, which shows remarkable few-shot performance for visual question answering. Subsequently, GPT-4~\cite{gpt4}, and LLaVA~\cite{llava} have brought in visual instruction tuning to improve the instruction-following ability of VLMs. Concurrently, models such as KOSMOS-2~\cite{kosmos2} have improved VLMs with visual grounding capabilities. The latest open-source models such as Intern-VL~\cite{internvl,internvl15} and DragonFly~\cite{dragonfly} have reached comparable performance to closed-source solutions like GPT-4~\cite{gpt4} in a few amount of parameters offering more scalable and customizable solutions for a large scale adoption. 

\subsection{Machine Learning for game testing}
\label{subsec:mldltesting}

Software product testing generally falls into two categories: manual testing and automated testing. Manual testing, particularly end-to-end play testing, remains the primary technique used by game developers. However, this method is often considered time-consuming and costly within the industry~\cite{surveyCyberpunk}. While manual testing provides valuable feedback that enhances software quality, there has been a significant shift towards applying existing automated testing techniques or developing new ones specifically for gaming.
This shift aims to provide developers with tools to evaluate software quality continuously and on time to provide a better experience for end users. By creating surrogates of human perception, it is possible to integrate testing into the entire software lifecycle instead of relying solely on predetermined testing phases with human testers.

Machine learning and deep learning techniques have been pivotal in this evolution. Gameplay was proposed as a source to investigate bugs in games~\cite{gameplayBug}, although the investigation was limited to video metadata and simple machine learning and deep learning solutions. Reinforcement learning and inverse reinforcement learning were also employed to create agents for defect detection learning from human tester~\cite{agentsHumanlike}. Frameworks such as RiverGame are built to automate various aspects of video game testing~\cite{frameworkGameTesting}, analyzing visual and sound aspects and using a behavior-driven development methodology for test specifications. 

The latest advancements, such as Large Language Models (LLMs), have been explored for detecting bugs in video games by interpreting textual descriptions of game events~\cite{llmBugText}. The effectiveness of Vision Language Models has been proven in Graphic User Interface (GUI) testing, as shown in VisionDroid~\cite{mllmGuiTesting} and CogAgent~\cite{vlmCogagent}. In the game testing field, only GlitchBench evaluates their effectiveness~\cite{glitchBench} to identify game glitches from Reddit images. 

Although many solutions have investigated game testing from different perspectives, to our knowledge, no solution has been proposed to investigate the effectiveness of game tutorials, which are critical to the success of a game in the early stages of gameplay. Furthermore, the capabilities of vision language models in this domain have not yet been sufficiently explored. This paper proposes to fill these gaps and pay more attention to these two underestimated aspects.

\section{Dataset}
\label{sec:dataset}
Publicly available games lack annotations about the developers' intention for specific choices in the tutorials, so we selected a game for which we can obtain such information from the developing team to provide an effective way to measure if the visual and textual components in the game effectively convey the intentions. The chosen game is But They Are Cats\footnote{\url{https://thefellowshipofthebox.itch.io/but-they-are-cats}}%
.

\subsection{Game and Tutorials}
The game is a tower defense with some extra mechanics explained in the four tutorials. This section summarizes the game's rules, describing what each tutorial wants to explain to players.

\subsubsection{Tutorial 1}
The first tutorial introduces the player to basic mechanics, which can be summarized as follows:
(1) The player has to place cats-in-boxes (the turrets) to defend the cheese from rats (enemies); (2) There are waves of enemies interleaved by some breaks; (3) If a rat reaches the cheese, the player can still save the cheese if he/she manages to destroy the rat before it can reach the base; (4) If a rat holding cheese is destroyed, the dropped cheese will slowly move toward its original place before being considered safe; (5) The level reward depends on the number of pieces of cheese rescued plus a base reward.

\subsubsection{Tutorial 2}
The second tutorial introduces the player to the behavior of the cats and can be summarized as follows: 
(1) Events can distract cats during the break between waves. As a result, cats will exit their boxes; (2) Player can attract cats to desired positions using different tools available; (3) Each cat has a favorite tool that speeds up the entire process.

\subsubsection{Tutorial 3}
The third tutorial introduces the player to the differences between types of cats and can be summarized as follows: (1) Different types of cats are available. The second type of cat is presented. It is faster and does less damage than the previous one; (2) There exist different types of rats; (3) Each type of cat reacts differently depending on distractions and tools used by the player.

\subsubsection{Tutorial 4}
The fourth tutorial introduces the player to the interactions with the environment and can be summarized as follows: (1) It is possible to move the camera around in the level; (2) The third type of cat is presented. It is slower than the others, but it can attack more rats at once; (3) Cats can jump on furniture; (4) Cats can break certain items placed around the level and alter the rats' path; (5) Each broken item around the level negatively affects the final reward; (6) It is possible to find special items in the levels.

\subsection{Data and Annotation}
\label{subsec:data_annotation}
\begin{figure}[htb]
    \centering
    \subfloat[Playable mode]{\includegraphics[width=0.45\linewidth]{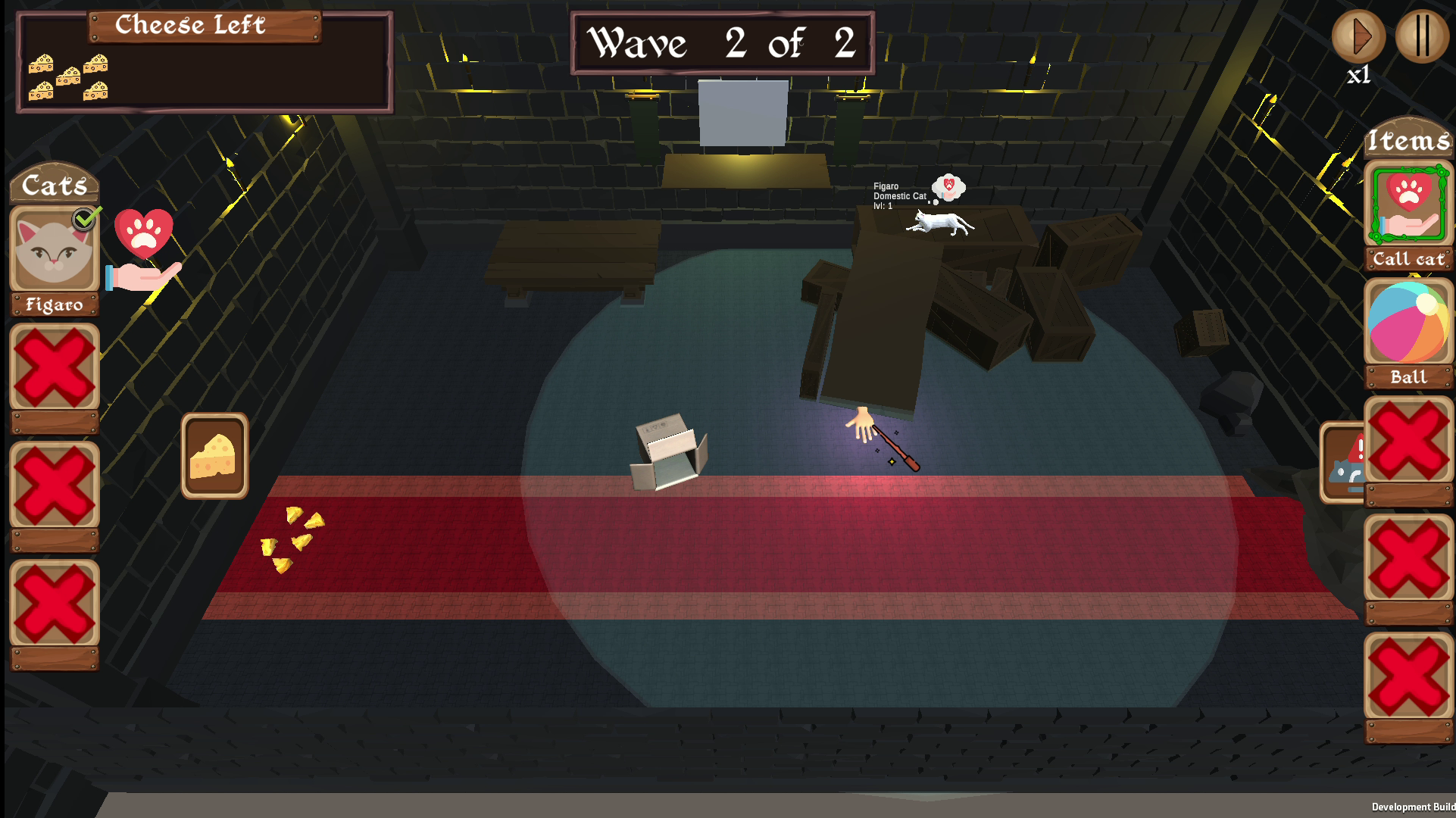}}
    \hfill
    \subfloat[Descriptive mode]{\includegraphics[width=0.45\linewidth]{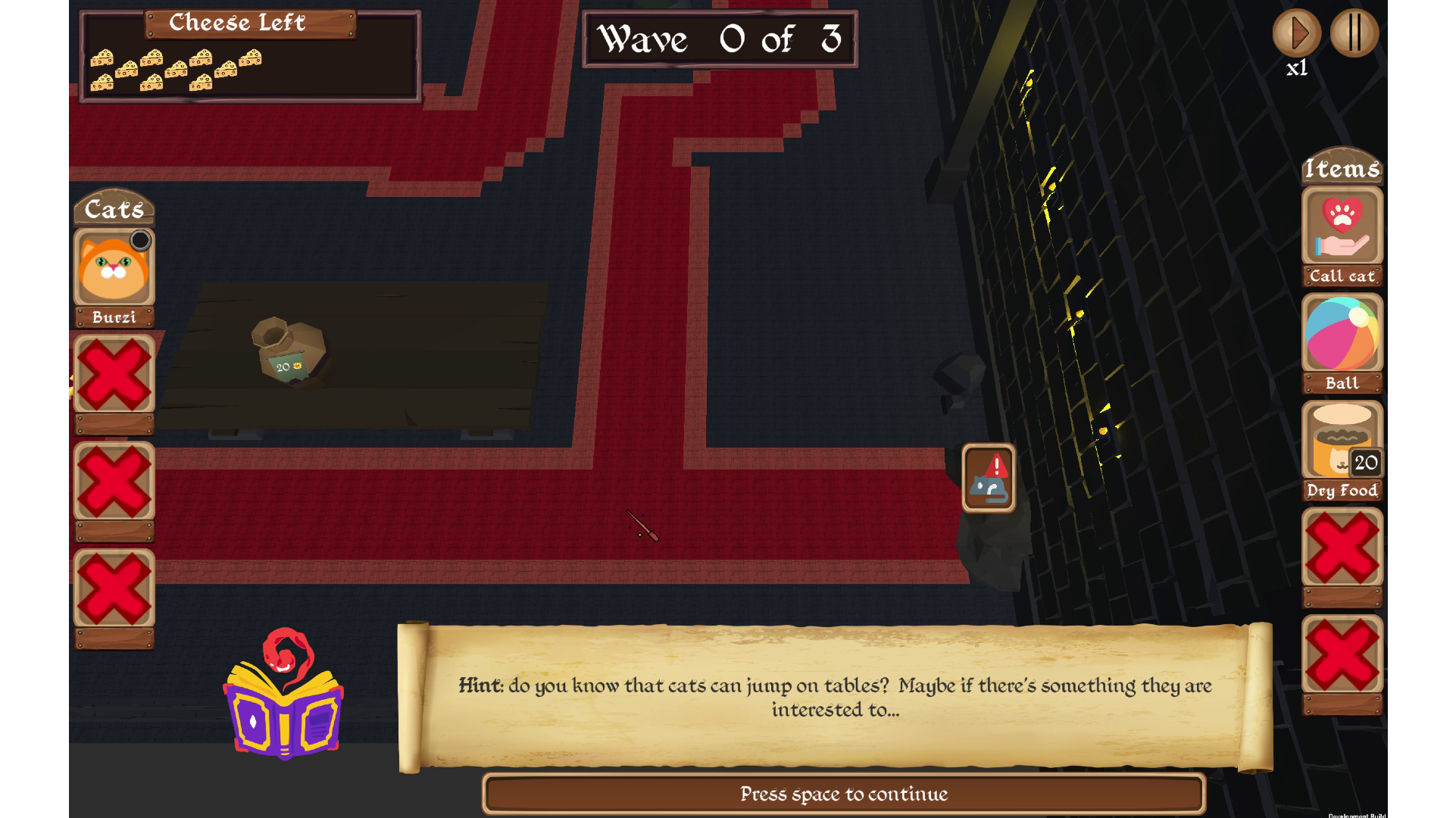}}
    \caption{Example of the two modes of the video game tutorial. }
    \label{fig:example_tutorial}
\end{figure}

Every video game tutorial alternates between two modes: a descriptive mode, which introduces the game mechanics to the player, and a playable mode, which requires player interaction as shown in~\Cref{fig:example_tutorial}. In contrast to automated testing methods that use game agents, this work focuses on analyzing the presentation part of each tutorial. We aim to determine whether visual language models (VLMs) can effectively substitute human perception to assess the comprehensibility of the visual and textual information presented to the user.

\begin{figure}[htb]
    \centering
    \includegraphics[width=\linewidth]{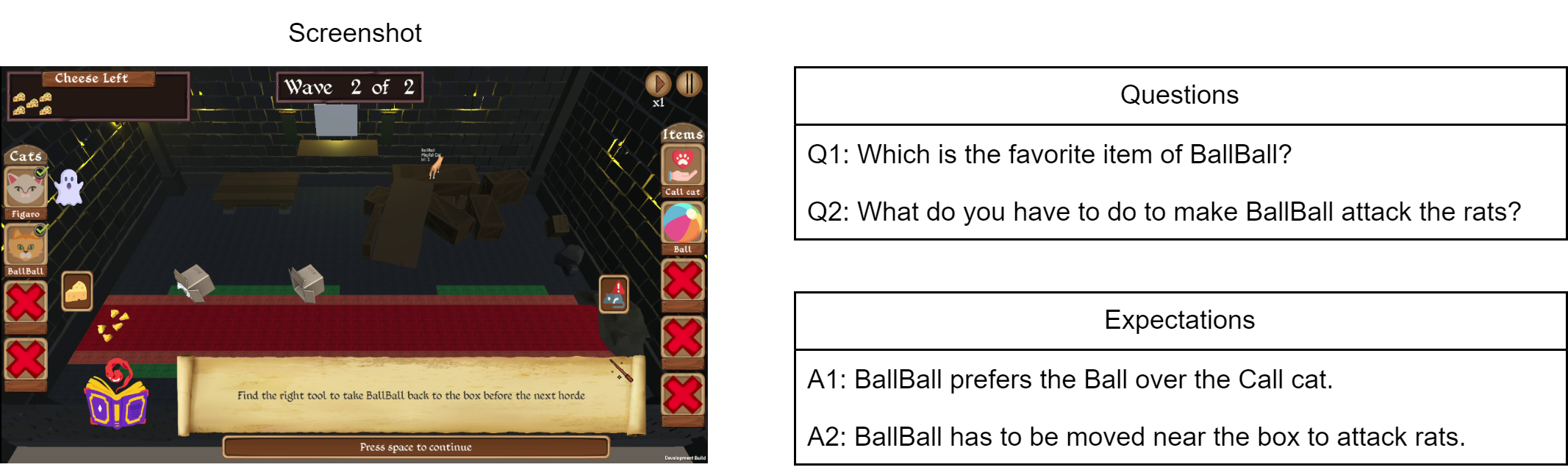}
    \caption{Example annotation. For each frame, we provide a list of questions and answers related to the frame.}
    \label{fig:example_annotation}
\end{figure}

To this end, our technique involves extracting significant frames from each recorded video of the tutorial. Each frame is labeled with questions that a developer would ask a human tester to determine if the purpose of that part of the tutorial is clear, along with the expected answers (ground truth) that demonstrate understanding.~\Cref{fig:example_annotation} shows one labeled frame. This process was repeated for frames from two different versions of the same game: %
one from an older development iteration (namely version P), and the latest released version (namely version L)

This approach allows us to test whether the improvements introduced in the latest version correspond to the quality assessment results obtained through our analysis. It is important to note that the game is a prototype with room for improvement, so even the scores obtained for the latest version are not for a final, polished game.

The answers are direct and the outputs generated by the VLM must be concise. This is enforced to avoid lower evaluation scores against the original developer's expected answer due to a strong mismatch of the lexicon or the semantic meaning of the phrases, since the purpose is to understand if specific concepts and rules about the game are sufficiently clear and understandable.
\subsubsection{Quality assessment technique}
Our objective is to create a tool that helps developers evaluate the quality of a game, one section at a time, following a methodology conceptually similar to writing automated tests for their codebase. The choice to label each frame with one or more question-answer pairs stems from the basic method of writing functional tests in the field of software engineering. The classical framework involves defining a test case and constructing the test following the AAA pattern (Arrange-Act-Assert) or the equivalent Given-When-Then in Behavior-Driven Development (BDD)~\cite{aaaUnitTest}. 

In our context, the question represents the system-under-test, the frame serves as the test context, and the answer is the expected value. In our case, the Arrange step involves defining the input and the objective: selecting the scene to test, extracting the frame, and identifying the detail to be assessed for comprehension by a human player, which translates to the question. The Act step involves submitting the question and the frame to the VLM and collecting the answers, which in the testing framework is often referred to as the \textit{actual value}. 

\begin{figure}[htb]
    \centering
    \includegraphics[width=\linewidth]{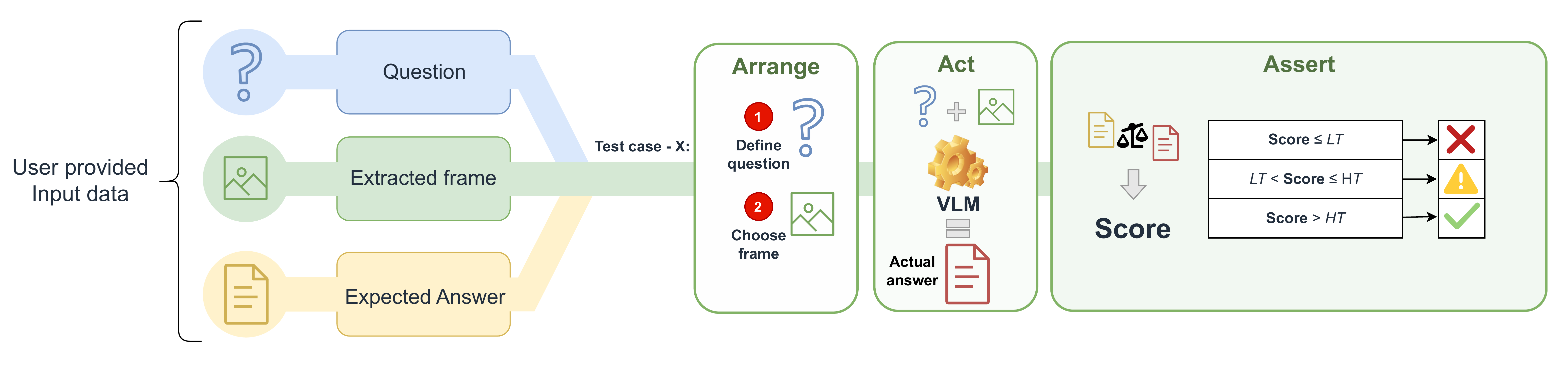}
    \caption{Matching of the proposed method with AAA pattern. The Arrange defines the question-frame couple to make a test on. The Act produces the Actual Answer given the text-image pair using a VLM. The Assert computes a score evaluating the Expected against the Actual Answer. This score is then compared to Lower and High Thresholds (LT, HT) to indicate whether the test is passed, requires revision, or failed.}
    \label{fig:aaaMethod}
\end{figure}

The assertion step evaluates the correspondence between the actual and expected values (the ground truth provided by the developer). This process determines whether the test passes or fails. As detailed in~\Cref{subsec:metrics}, we have established thresholds based on chosen metrics to provide developers with an indication of the test outcome. In our quality assessment approach, the developer receives feedback on whether the frame is acceptable, needs revision, or is rejected. The matching between AAA and our proposed method is illustrated in~\Cref{fig:aaaMethod}.

\section{Methodology}
\label{sec:methodology}
In this section, we describe the models employed for the study, the metrics adopted in the experimental evaluation, and how they can be interpreted to evaluate the quality of the videogame's tutorial.

\subsection{Models}
In this section, we describe the VLMs employed for our study. GPT-4~\cite{gpt4} is the closed-source model that achieves state-of-the-art performance on many benchmarks. More in detail, we evaluated the GPT-4o model in the tutorial evaluation task. Instead, regarding open models, we considered the InternVL family and DragonFly model. The first one~\cite{internvl,internvl15} is a family of open models that proved good performance on many benchmarks (comparable to GPT-4). Different sizes of InternVL models are available, ranging from 2B to 108B parameters. Instead, DragonFly~\cite{dragonfly} is a recent small open-source model that supports multiple image resolutions.

\subsubsection{GPT-4}
GPT-4o~\cite{gpt4} is a large-scale, multimodal model developed by OpenAI, capable of processing both text and image inputs to generate text outputs. Based on the Transformer architecture, it was pre-trained on a vast corpus of data and fine-tuned using Reinforcement Learning from Human Feedback (RLHF). GPT-4 demonstrates human-level performance on several professional and academic benchmarks, including scoring in the top 10\% on a simulated bar exam. It excels in various NLP tasks, surpassing previous models and state-of-the-art systems in multiple languages.

\subsubsection{InternVL}
InternVL~\cite{internvl} is a large-scale vision-language model designed to integrate a robust vision encoder with a large language model, utilizing a custom vision transformer, InternViT-6B, and a language middleware, QLLaMA. Its training strategy is composed of vision-language contrastive training, generative training, and supervised fine-tuning. InternVL 1.5~\cite{internvl15} enhances the original with continuous pre-training and dynamic resolution achieving state-of-the-art results across benchmarks and available in different sizes. The latest, InternVL 2.0, introduces progressive alignment training, supports multiple input modalities and multitask outputs, and achieves superior performance in multimodal tasks, available in sizes from 2B to 108B parameters.

\subsubsection{DragonFly}
DragonFly~\cite{dragonfly} is designed to enhance fine-grained visual understanding and reasoning, particularly for high-resolution images. It employs a multi-resolution visual encoding strategy, processing images at low, medium, and high resolutions and a zoom-in patch selection method that focuses on the most informative high-resolution sub-images. Visual tokens from these images are projected into the language space and combined with text tokens for input into a large language model. This architecture allows Dragonfly to efficiently handle high-resolution images, making it especially effective for tasks requiring detailed image analysis. Dragonfly-Med, the biomedical version, demonstrates state-of-the-art performance on various medical image benchmarks, validating its design.

All models' characteristics and their respective scores in multimodal understanding benchmarks are summarized in~\Cref{tab:model_comparison}.

\begin{table}[htb]
    \centering
    \caption{Models' comparison on MMMU~\cite{mmmu_bench} and MMBench~\cite{MMBench} benchmarks according to Open VLM Leaderboard~\cite{openvlmleaderboard} or respective paper.}
    \label{tab:model_comparison}
    \begin{tabular}{l|c|ll|cc}
\toprule
\textbf{Model}               & \textbf{Open}   & \textbf{Language Model} & \textbf{Vision Model}     & \textbf{MMMU} & \textbf{MMBench} \\ \midrule
GPT-4o \cite{gpt4}             & \xmark & -              & -                & \textbf{69.2} & \textbf{82.2}    \\
InternVL2-26B \cite{internvl}      & \cmark & InternLM2-20B  & InternViT-6B     & 55.2 & 81.2    \\
InternVL2-8B \cite{internvl}       & \cmark & InternLM2-7B   & InternViT-300M   & 51.2 & 79.4    \\
InternVL2-4B \cite{internvl}       & \cmark & Phi-3-mini \cite{phi3}    & InternViT-300M   & 48.3 & 73.6    \\
InternVL1.5-26B \cite{internvl15}        & \cmark & InternLM2-20B  & InternViT-6B     & 46.8 & 79.7    \\
InternVL1.5-4B \cite{internvl15}     & \cmark & Phi-3-mini \cite{phi3}     & InternViT-300M   & 45.1 & 69.7    \\
DragonFly \cite{dragonfly} & \cmark & Llama3-8B \cite{llama3}      & CLIP \cite{clip} & 36.2 & -       \\ \bottomrule
\end{tabular}
\end{table}

\subsection{Tutorial evaluation}
As introduced in~\Cref{subsec:data_annotation}, we extracted from the four tutorials different frames depicting and introducing to the players several game mechanisms. Data acquisition was conducted on two different versions of the videogame tutorial: its \textit{latest} release, namely L, and the \textit{previous} deployment, which we identify as P. Each tutorial in our dataset is characterized by several screenshot acquisitions, and each capture is associated with one or more questions and their corresponding answers. Each pair question/answer was formulated by the developers and entails details about the game's mechanics, which are described in the tutorial phase. The changes that were made after revising the video game led to differences in tutorial details, so the number of acquisitions between P and L is different.

All the considered VLMs in this paper were evaluated in two different experimental configurations:
\begin{itemize}
    \item with history: at every prompt, the model received as input the entirety of chat history for the selected tutorial, in conjunction with the new frame to be evaluated as well as all its related questions, each being appropriately numbered
    \item without history: at every prompt, the VLM received as input the new frame and all its related questions without any previous information
\end{itemize}
The former configurations were proposed to evaluate the performance difference in scene understanding, also considering game mechanics that were presented in previous images, in contrast with the sole information contained within the new frame. 

Additionally, as an initial prompt, we provided the following phrase to the system: \textit{"You are a gamer. You are playing a game tutorial. I will provide you some screenshots of the tutorial. Answer the questions related to the screenshot. Be concise and direct."}. Such prompt was provided at the beginning of every chat session, i.e., the "with history" configuration provided such phrase as input only for the first question, whereas the "without history" configuration appended it at the beginning of every question. 

The evaluation procedure split all the numbered answers to match the corresponding question. Metric computation was performed on a per-question basis, i.e., we evaluated each response separately with the expected answer. This solution permits the identification of the unclear aspects of each frame.

\subsection{Metrics}
\label{subsec:metrics}
To understand if the visual explanations are sufficiently clear and to provide automatic evaluation of the output quality, we need a quantitative measure of the results. We rely on established metrics for summary evaluation: ROUGE~\cite{rouge} and BERT-Score~\cite{bert-score}.

ROUGE-N~\cite{rouge} evaluates the syntactical overlap between N-grams (contiguous sequences of N items from a given text) of the prediction and of the ground truth. In this study, we employed ROUGE-1 (the overlap of unigrams), ROUGE-2 (the overlap of bigrams), and ROUGE-L (that measures the longest common subsequence) as ROUGE-N metrics.

Instead, BERT-Score~\cite{bert-score} relies on encoder-only language models to measure the semantical overlap. For each token in the generated text, BERT-Score computes the cosine similarity with every token in the reference text. This results in a similarity matrix where each element represents the semantic similarity between a token in the generated text and a token in the reference text. The precision and recall are calculated to measure how many tokens are in common. The F1-Score aggregates the two metrics.

\section{Experimental results}
\label{sec:results}
This section presents experimental settings together with qualitative and quantitative performances.

\subsection{Experimental Settings}
All experiments were conducted on a workstation with Intel Core i9-10980XE, 128GB of RAM and an NVIDIA RTX A6000 GPU with 48GB of VRAM. 
All open-source models were employed with zero temperature for determinism. The images were resized at 1920x1080 resolution and preprocessed according to each model requirements. BERT-Score was calculated employing all-mpnet-base-v2 of Sentence Transformers~\cite{sentence-bert} trained for the sentence embedding task. The GPT-4 version employed is GPT-4o accessed on July 2024.

\subsection{Comparative Results between Versions}
In this section, we compare the scores achieved by all the models on common questions between the two tutorial versions (\textit{latest} and \textit{previous}). The results are shown in Table~\ref{tab:results} for both the "with history" and "without history" configurations. It can be observed that, for all the considered models, metrics, and configurations, the latest version demonstrates higher scores and improvements compared to the previous version. This supports the fact that our proposed approach is able to capture improvements over the revised version of the game, ranging from +1.61\% to +46.75\% with history information and ranging from +2.10\% and +45.21\% without any previous interaction with the model. In both cases, the most significant increase was observed in ROUGE-2 metric, independently from the model. These results also suggest the historical information is not necessarily relevant for evaluating the frames since we can see differences in any case.

\begin{table}[htb]
\caption{Results obtained for the previous version (P) and the last one (L) by tested models for common questions}
\label{tab:results}
\centering
\begin{tabular}{@{}c|l|cc|cc|cc|cc@{}}
\toprule
            &    & \multicolumn{2}{c|}{ROUGE-1} & \multicolumn{2}{c|}{ROUGE-2} & \multicolumn{2}{c|}{ROUGE-L} & \multicolumn{2}{c}{BERT-SCORE} \\ \midrule
& Model           & P        & L                 & P        & L                 & P        & L                 & P         & L                  \\ \midrule
\multirow{7}{*}{\rotatebox[]{90}{with hystory}} & GPT-4o            & 0.4125   & \textbf{0.4998}   & 0.1844   & \textbf{0.2706}   & 0.3664   & \textbf{0.4361}   & 0.6505    & \textbf{0.6973}    \\
& InternVL2-26B   & 0.4136   & \textbf{0.4540}    & 0.1775   & \textbf{0.2488}   & 0.3563   & \textbf{0.4080}    & 0.6480     & \textbf{0.6679}    \\
& InternVL2-8B    & 0.4338   & \textbf{0.4466}   & 0.1883   & \textbf{0.2479}   & 0.3646   & \textbf{0.3872}   & 0.6641    & \textbf{0.6748}    \\
& InternVL2-4B    & 0.4352   & \textbf{0.4485}   & 0.1957   & \textbf{0.2383}   & 0.383    & \textbf{0.4025}   & 0.6504    & \textbf{0.6771}    \\
& InternVL1.5-26B & 0.3811   & \textbf{0.4214}   & 0.1659   & \textbf{0.2156}   & 0.3168   & \textbf{0.3676}   & 0.6390     & \textbf{0.6563}    \\
& InternVL1.5-4B  & 0.4555   & \textbf{0.4780}    & 0.2119   & \textbf{0.2459}   & 0.3841   & \textbf{0.4173}   & 0.6544    & \textbf{0.6848}    \\
& DragonFly       & 0.2764   & \textbf{0.3073}   & 0.1261   & \textbf{0.1565}   & 0.2459   & \textbf{0.2788}   & 0.5897    & \textbf{0.6369}    \\ \midrule
\multirow{6}{*}{\rotatebox[]{90}{without hystory}} & InternVL2-26B   & 0.402   & \textbf{0.4187}    & 0.1862   & \textbf{0.2311}   & 0.3525   & \textbf{0.3751}    & 0.6424     & \textbf{0.6578}    \\
& InternVL2-8B    & 0.3862   & \textbf{0.4273}   & 0.1617   & \textbf{0.2348}   & 0.3274  & \textbf{0.3744}   & 0.6442    & \textbf{0.6633}    \\
& InternVL2-4B    & 0.4381   & \textbf{0.4819}   & 0.2016   & \textbf{0.2527}   & 0.3783    & \textbf{0.4169}   & 0.6611    & \textbf{0.6848}    \\
& InternVL1.5-26B & 0.3566   & \textbf{0.3706}   & 0.1563   & \textbf{0.1889}   & 0.3036   & \textbf{0.3236}   & 0.6367     & \textbf{0.6625}    \\
& InternVL1.5-4B  & 0.4398   & \textbf{0.4620}    & 0.204   & \textbf{0.2410}   & 0.3748   & \textbf{0.3886}   & 0.6482    & \textbf{0.6618}    \\  
& DragonFly       & 0.3189   & \textbf{0.3646}   & 0.1394   & \textbf{0.1954}   & 0.2845   & \textbf{0.3275}   & 0.5985    & \textbf{0.6146}    \\ \bottomrule %
\end{tabular}
\end{table}

\subsection{Models consistency}
\begin{figure}[htb]
    \begin{minipage}{0.45\linewidth}
        \centering
        \includegraphics[width=\linewidth]{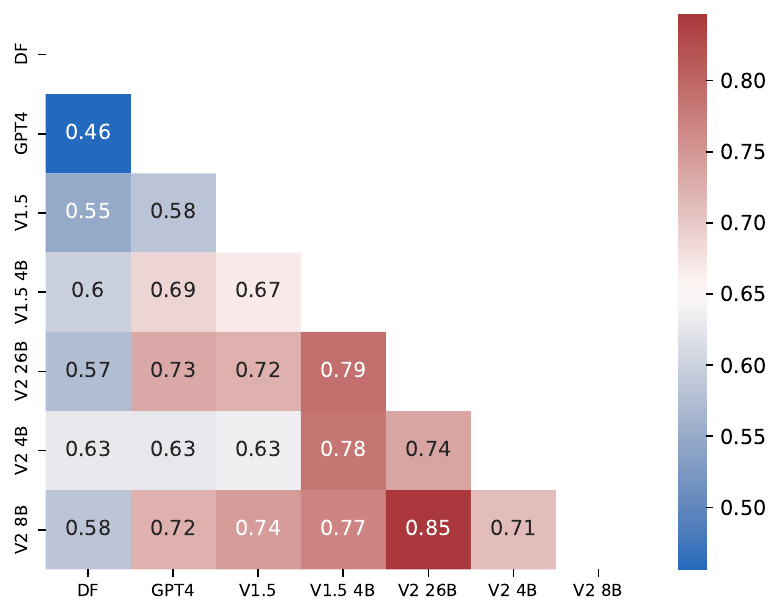}
        \captionof{figure}{Mean Spearman Correlation across metrics by model. \textit{DF} is DragonFly, and \textit{VX} indicates the InternVL version and eventually the size.}
        \label{fig:correlation}
    \end{minipage}
    \hfill
    \begin{minipage}{0.45\linewidth}
        \centering
        \includegraphics[width=\linewidth]{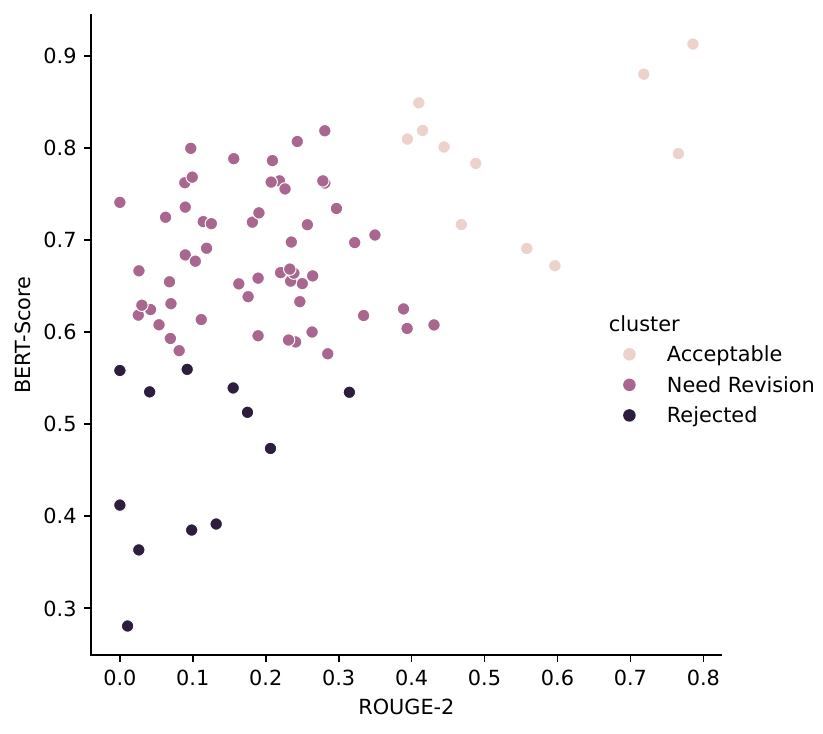}
        \captionof{figure}{Clusters of quality level according to ROUGE-2 and BERT-Score}
        \label{fig:cluster_by_quality}
    \end{minipage}
\end{figure}

Given that we have tested different models, it might be beneficial to consider whether it would be possible to employ one model in place of another, depending on whether they are able to identify good and bad samples consistently. We compute the Spearman Rank correlation in~\Cref{fig:correlation} to understand the model agreement. The correlations are all medium-high and the lowest are achieved by the worst model in performance: DragonFly. The highest agreement is shown by InternVL V2 8B and 26B. This fact proves that it is not necessary to use the larger version because similar results can be obtained using the smaller version, which is more portable. They both show correlations higher than 0.7 with GPT-4, representing good open-source alternatives.

\subsection{Analysis of Tutorials Clarity}
In this section, we evaluate each tutorial from the \textit{latest} version to understand if there could be any differences between their clarity level and where there could be room for improvements. We compute the mean metrics over tutorials and models. In~\Cref{tab:tutorial_results}, we can see the differences between tutorials and metrics are consistent, i.e. the ROUGEs ranks coincide with the BERT-Score rank. The first tutorial proves to be, in general, the clearest one, while the third tutorial is the unclearest one. From this perspective, we can conclude the third tutorial needs attention due to possible mistakes or unclear passages in the teaching and learning process.

\begin{table}[htb]
    \centering
    \caption{The mean results by tutorial and models. The best one is in bold, and the worst one is underlined.}
    \label{tab:tutorial_results}
    \begin{tabular}{l|cccc}
    \toprule
               & ROUGE-1 & ROUGE-2 & ROUGE-L & BERT-SCORE \\ \midrule
    Tutorial 1 & \textbf{0.4535} & \textbf{0.26} & \textbf{0.4004} & \textbf{0.6893} \\
    Tutorial 2 & 0.4126 & 0.1882 & 0.3371 & 0.6546 \\
    Tutorial 3 & \underline{0.3261} & \underline{0.1226} & \underline{0.2779} & \underline{0.6345} \\
    Tutorial 4 & 0.4182 & 0.237	& 0.3834 & 0.6398 \\
\end{tabular}
\end{table}

\subsection{Automated Testing and Qualitative Analysis}

\begin{figure}[htb!]
    \centering
    \subfloat[Example 1]{
        \begin{tabular}{l|p{0.8\linewidth}}
            \toprule
            \textbf{Question} & What is the Treasure? \\ \hline
            \textbf{Expectation} & The cheese to protect is the Treasure. \\ \midrule
            \textbf{Answer} & The Treasure is an item located on the left side of the screen, indicated by the label 'Treasure'. \\\hline
            \textbf{Score} & BS: 0.3871, R2: 0.087 \\\bottomrule
            \multicolumn{2}{c}{} \\
            \multicolumn{2}{c}{\includegraphics[width=0.6\linewidth]{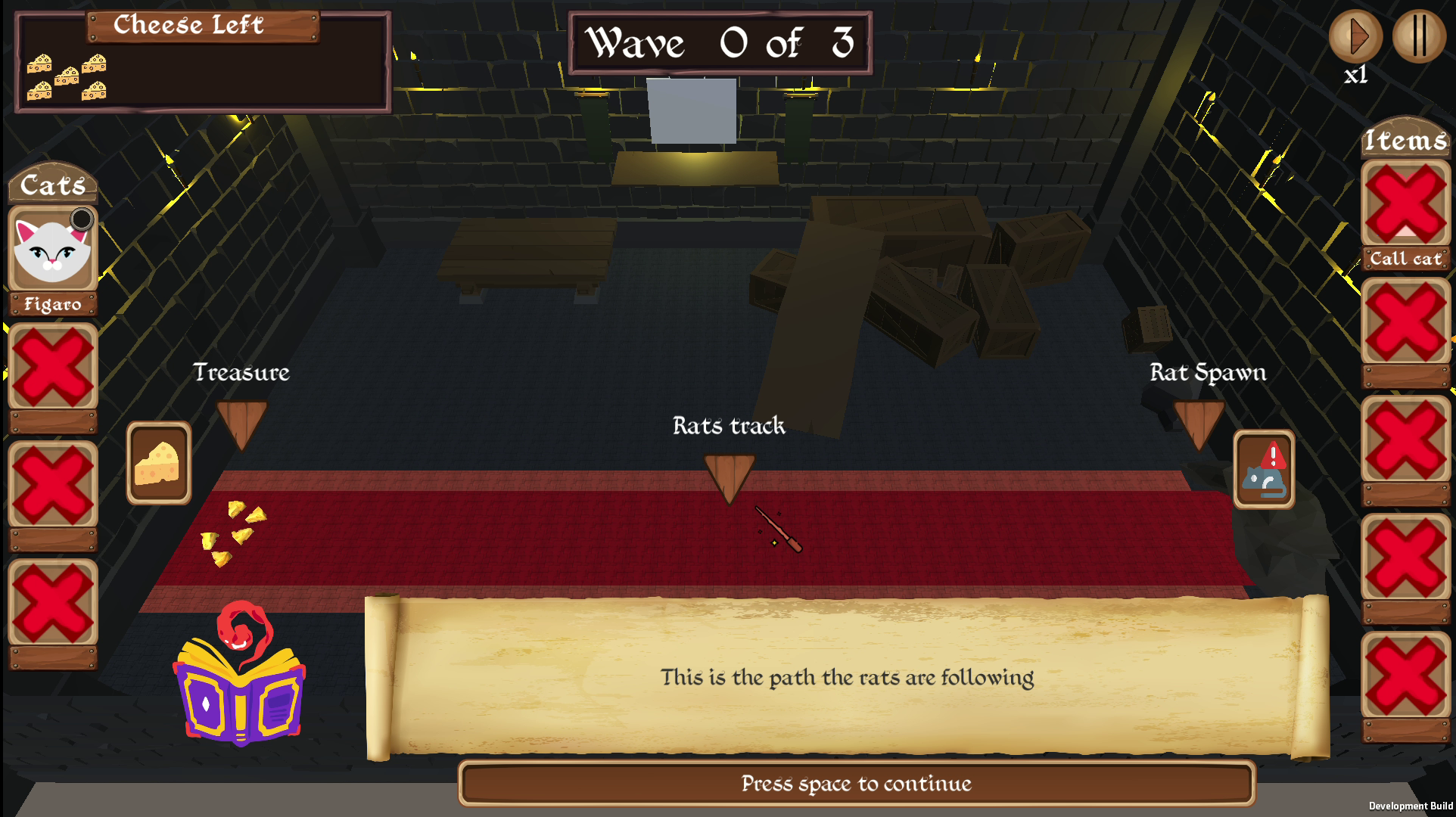}}
        \end{tabular}
    } \\\vspace{5mm}
    \subfloat[Example 2]{
        \begin{tabular}{l|p{0.8\linewidth}}
            \toprule
            \textbf{Question} & How can you know what is distracting a cat? \\\hline 
            \textbf{Expectation} & The indicator near the cat in the Cats panel indicates what distracts the cat. \\ \midrule
            \textbf{Answer} & Distractions may be indicated by visual or audio cues in the game environment, but the specific indicator is not shown in the screenshot. \\\hline
            \textbf{Score} & BS: 0.4898, R2: 0.0571 \\\bottomrule
            \multicolumn{2}{c}{} \\
            \multicolumn{2}{c}{\includegraphics[width=0.6\linewidth]{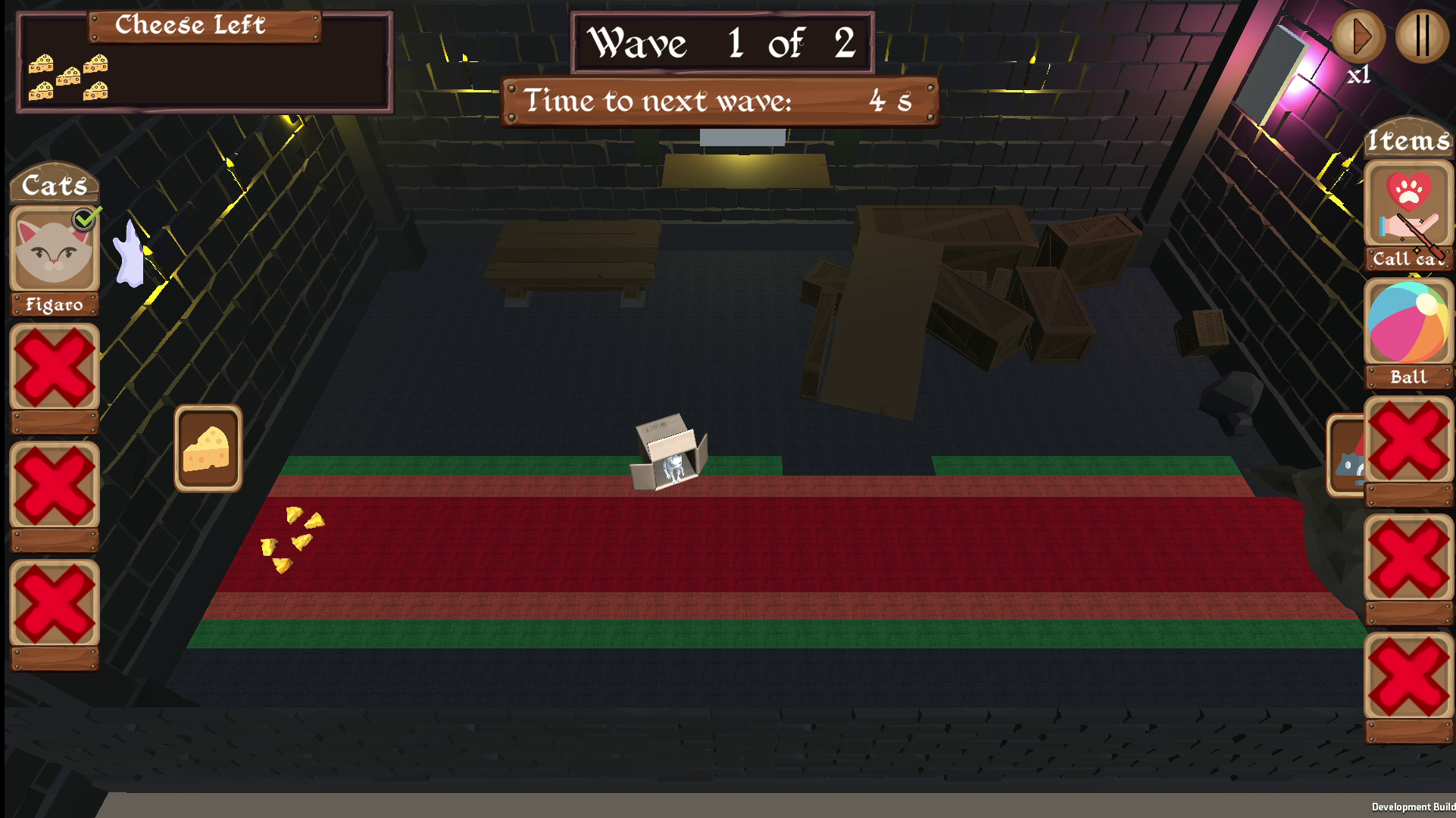}}
        \end{tabular}
    }
    \caption{Example of bad samples using the thresholds from clustering. Each example show the reference frame, the question, the expected answer (Expectation), the predicted answer (Answer) and the relative ROUGE-2 (R2) and BERT-Score (BS).}
    \label{fig:bad_samples}
\end{figure}

It is possible to adopt many strategies to select the frames that require more attention due to clarity and should be revised by developers to improve the quality of the tutorial. 
To provide an automatic evaluation, we could consider three levels of attention for each answer: correct, requires attention, and wrong. Consequentially the related frame of the game could be considered acceptable, needing revision, or rejected, since its clarity is not sufficient.

Since we do not know apriori a good threshold to partition each frame in the three classes, we apply K-Means clustering using ROUGE-2 and BERT-Score of each model as features to create three clusters of similar samples. ROUGE metrics are highly correlated with each other, so we take only ROUGE-2 which is commonly employed in benchmarking. After such step, we obtain three clusters for the three different aforementioned quality levels as shown in~\Cref{fig:cluster_by_quality}. One simple solution to classify a frame based on its score is to apply K-NN (since they cannot be easily represented by linear equations ensuring a certain robustness).

An alternative approach, which provides simpler boundaries, is applying automatic thresholding algorithms like the multi-otsu one, which provides thresholds based on histogram analysis~\cite{Liao2001-dz}. In this case, we get $BS \leq 0.5149$, $R2 \leq 0.1785$ for low quality, and $BS > 0.7215$, $R2 > 0.4496$  for high quality ones. It's important to note that this method can only consider a single feature at a time, so it may not consider the relationship between R2 and BS in order to select the most suitable threshold.

In~\Cref{fig:bad_samples}, we show two samples that are below the provided thresholds with the answers provided by GPT-4. In the first case, the model does not understand the visual cue, probably due to a missing explanation. In the second case, the other visual cue is not clearly explained and could be unnoticed by both players and models for the same reason as before. In this way, we can understand easily were we should pay attention and which section could be improved.

\section{Conclusion}
\label{sec:conclusion}

This study presents an innovative approach to evaluating the quality of video game tutorials by leveraging Vision-Language Models (VLMs). Our research demonstrates that VLMs can effectively simulate human perception to assess the clarity and comprehensibility of game tutorials. By analyzing frames from tutorial videos and comparing VLM-generated responses to expected answers, we can identify confusing or problematic scenes, providing valuable feedback to developers.

Our experimental results show that the latest version of the game tutorials, which had undergone refinements, outperformed the previous version in terms of clarity and comprehensibility. This indicates that our automated approach can reliably detect improvements and issues within game tutorials, making it a useful tool for developers.

Additionally, the study reveals that historical information is not always necessary for evaluating frames, as both configurations with and without history showed improvements. This flexibility can help the evaluation process, reducing the need for a large context and using this approach for different lengths of tutorials.
The clustering technique and Otsu's method used to classify frames into different quality levels proved practical and effective in offering a method for developers to focus on specific areas needing improvement. 
However, one limitation is ensuring that the model's perception aligns with that of actual players, which could vary based on experience and familiarity with the game.

Future work will aim to compare VLM assessments with human evaluations more comprehensively and extend the methodology to detect other issues such as verbal harassment, bugs, and non-inclusive content. This will enhance the robustness of our approach, providing a more thorough assessment of game tutorial quality and overall player experience.

\section*{Acknowledgements}
Special thanks go to Bincoletto Alessio, Bodnarescul Paolo Stefanut, and Riola Mattia for their collaboration in realizing the game itself.

\bibliographystyle{splncs04}
\bibliography{main}
\end{document}